# Overarching Computation Model (OCM)

Henok Ghebrechristos[1]
University of Colorado-Denver
henok.ghebrechristos@ucdenver.edu

Drew Miller
rbitrator@live.com

## Abstract

Existing models of computation, such as a Turing machine (hereafter, TM), do not consider the agent involved in interpreting the outcome of the computation. We argue that a TM, or any other computation model, has no significance if its output is not interpreted by some agent. Furthermore, we argue that including the interpreter in the model definition sheds light on some of the difficult problems faced in computation and mathematics. We provide an analytic process framework to address this limitation. The framework can be overlaid on existing concepts of computation to address many practical and philosophical concerns such as the P vs NP problem. In addition, we argue that the P vs NP problem is reminiscent of existing computation model which doesn't account for the person that initiates the computation and interprets the intermediate and final output. We utilize the observation that *deterministic computational procedures lack fundamental capacity to fully simulate their non-deterministic variant* to conclude that the set NP cannot be fully contained in P. Deterministic procedure can approximate non-deterministic variant to some degree. However, the logical implication of the fundamental differences between determinism and non-determinism is that equivalence of the two classes is impossible to establish.

## 1 INTRODUCTION

The notion of computation has existed in some form since the dawn of the mankind. Although a concise definition of the term computation is hard to come by, many describe it as the act of decision making, within some finite decision space, in order to achieve some end goal. Its theoretical foundations started with the Turing machine (Anil Maheshwari and Michiel Smid, 2017), a theoretical computational model. *Turing based his model on how he perceived mathematicians think.* As digital computers were developed in the 40's and 50's, the Turing machine proved itself as the right theoretical model for computation. In the first half of the 20th century, the notion of computation was made much more precise than the hitherto informal notion (Fortnow and Homer, 2002). Many different equivalent models of computation were discovered —Turing machines, lambda calculus (Rojas, 2015), cellular automata (Linz, 2012), pointer machines (Ben-Amram, 1998), Conway's Game of life (Yu and Reevesman, 2015), etc. The computer's rapid adoption in society in the subsequent half decade brought computation into every aspect of modern life and made computational issues important in human endeavors. However, computation is not just a practical tool, but also a major scientific concept. Generalizing from models such as cellular automata, one can view many natural phenomena as analogous to computational processes. Below we discuss a framework derived from this analogy. The framework assumes *process* to be more fundamental than cellular automata or any other models and provides formalization for observable and synthesizable processes such as those that occur in computation. It's our conviction that Turing based computation models are inferior to analytic process models. When we base computation models on processes in general as oppose to only physically realizable processes, our understanding of both nature and the essence of digital computation is enhanced.

# 2 ANALYTIC PROCESS

**"All religions, arts and sciences are branches of the same tree."** Albert Einstein

A dictionary definition of the term process is a systematic series of actions directed to some end. In this text, the term process is used to describe an observed act or series of actions of transforming some input to some output within some decision space. It's important to emphasize that the term, process, is used as a *metaphor* because the traditional meaning captures what is discussed below. Three types of processes are considered to cover a wide-range of phenomenon that are encountered not only in computation but also in nature.

1. **Overarching Process (OP)** - an *environment* in which *transformation* of *input* is achieved by applying some *procedure* composed of a finite count of discrete *steps*.
2. **Analytic Process (AP)-** an *environment* in which transformation of some *input* to a <u>desired output</u> is achieved by applying some procedure composed of a finite count[1] of discrete steps.
3. **Computational Process (CP)** – an *environment* in which *transformation* of *digitized input data* to a desired *output* data specified a priori is achieved by applying some procedure composed of a finite count of discrete steps <u>which are carried out using a constrained well-defined mechanical device</u>.

Close attention is paid to the informal definition of process and thus highlights the aims of these definitions and our intentions to formally define a process framework. Analytic and computational processes are constrained versions of overarching process (OP), so one can consider the aim of any AP or CP considering OP. The informal definition of OP assumes all the components and instructions required to carry out the process are prespecified (provided) within the **environment**. Only when these components are put together and initiated does the process emerge. Notice, to speak of the process in this fashion, one must observe it to classify it as such. Hence an **observer** of some sort is integral part of the formalism. In addition, as stated in the definition, an OPs' concern is to ***transform an input*** (to a specific *output* or not) or rather the act of transforming input is an OP. Why is this important? Everything any human knows or does can be described as a transformation of input. For illustration, say a person wakes up in the morning and makes the bed. If we interpret this using OP, its equivalent to saying, 'transform the bedroom + unmade bed into the bedroom + a made bed'. Notice, all the components make up your environment: bed, room, bed sheets, person (the actor) are provided and the input state is <u>bedroom + unmade bed</u> and the output state is <u>bedroom + made bed</u>. Once the person starts making the bed to achieve the goal, an OP is carried out. To achieve the goal, decide to remove the pillows, then pull the bed sheet over the mattress, then place the pillows back. The *discrete steps* are; 1. remove the pillows, 2. pull the bed sheet over the matters and 3. place pillows back on the bed. Without these or similar steps, achieving the desired goal is impossible. It is important to note that the steps are closely tied to the goal and can only be recognized by the observer. However; each step is not, and need not be, part of the definition or formalism. It is an outcome of the process. Applying these steps in some order: first remove pillows, then pull the sheet, then place back the pillows, comprises the *procedure* (or algorithm in case of computation) that enables the goal to be achievable. Similar arguments can be made for any event that can be observed and synthesized. The ***observer*** can be any system that can record the steps.

---

[1] OPs are not necessarily results of construction but rather fundamental aspects of nature. They are not mere representations of computation as they relate to how human performs calculations or any tasks alike. They are models for natural and observable processes that involve discrete steps. This can be a person walking, natural cycle of day and night, a honey bee searching for nectar source, a computer running to solve a problem, a mathematician engaged in solving a problem etc.…

The differences between overarching and analytic process (AP) lie in the goal and start state of each process. OP considers a goal that is null. This is like a process that runs without any predefined desired goal, final state, or output. An OP also does not assume a known beginning or initial state. Any observer is capable in classifying a process with these constraints as an OP. For instance, consider a visual exercise. An observer stands on a river bank. The origins or endings of the river are unknown. The observer faces no challenge recognizing the river as a process, whereby the river possesses headwaters, the current river being observed came from something and somewhere. The river also has a destination. Similar arguments can be made for any event whose origins are humble and unknown. These types of processes, which are abundant in nature, are considered OP. AP on the other hand requires a start state and a goal state (desired goal). Because of the definition of AP, it is necessary for the **observer** to be or possess some sort of intelligence or a system which benefits from transformation of the input to the desired output. Yet another, even more constrained than AP, process that describes digital computations is termed computational process (CP). Below CP is discussed conceptually as it relates to a Turing machine.

### 2.1.1 Computation as a process

Let us elaborate CP using the concept of abstract computation. A model widely used to represent digital and theoretical computation is the Turing machine. Informally the Turing machine *(TM)* is a model that consists of a finite *state control, tape head*, and *tape* (Anil Maheshwari and Michiel Smid, 2017). These components enable it to carry out any computation making it capable of simulating any physically realistic computing device. The idea is that the action of TM at each instant is determined by the state of the finite state control together with the symbol on the tape the head is currently reading. Depending on the state and the symbol being scanned, the action that the machine is to perform may involve changing the state of the finite state control, changing the symbol on the tape square being scanned, and/or moving the tape head to the left or right. Once this action is performed, the machine will again have some state for its finite state control and will be reading some symbol on its tape, and the process continues. It is also give an opportunity to stop the computation and produce an output. In the most basic model there are two distinct special states of the finite state control: an *accept* state and a *reject* state. If the machine enters one of these two states, the computation immediately stops and accepts or rejects accordingly.

It is trivial to show a running TM, when observed, undergoes an analytic process. Hence the proposed framework not only encapsulates our existing understanding of computation but also extends its capacity to explain other *fundamental concerns* such as the P vs NP problem. It's important to note that a digital computation is an analytic process performed on mechanical device. Notice TM doesn't consider the agent involved in interpreting the outcome of the computation. *We argue that a Turing machine, or any other computation model, has no significance if its output is not interpreted by some agent. Furthermore, we argue that including the interpreter in the model sheds light on some of the difficult problems faced in computation and mathematics ((Aaronson, 2011) highlights few of these problems).* We propose an all-embracing abstract computation model that not only includes the mechanism for performing computation but also integrates an observer that records and interprets the steps and outputs. The observer records input states at any given computation stage along with the rules applied to transform the input. The recorded information can then be served as an input to the interpreter to assess the computation path taken to transform the initial input state to some output state.

## 2.2 OVERARCHING PROCESS

As discussed above, an OP captures all phenomenon that occur in nature. As a result, it is the mother of all other processes. OP describes the process of transforming some input to an output by performing well defined discrete operations. Formally an OP must be defined in terms of its individual components (Definition 2.1).

### 2.2.1 Definition 2.1: An overarching process (OP) is an 8-Tuple ⟨$I, O, G, Σ, Γ, Δ, Q_0, M$⟩ where,

- I – input data composed of finite symbols
- O – is an observer system
- M – a mechanism for applying transition rules
- G (I) – describes the outcome expected (can be null) - can also be encoded in O
- Σ – a finite set of symbols
- Γ – a structure for storing symbol (grid of cells) of the input
- Δ – a finite set of rules of transition
- $Q_0$ – starting configuration of I and the environment

Transformation of the input I is carried out using discrete operations (for example bit shifting if the input comprises of bits). The operations are applied with a mapping function Δ that takes a given state of the environment and returns the same environment (if the operation is no-operation) or a changed environment, if the operation enforces a change in state. A state change in this context can be defined as the opposite of same or no change. For instance, a change can be flipping a bit (from 0 to 1 in case of a bit comprised input), moving the head of a TM to the left etc. The environment defines a closed system whereby the process can run without external influence. For the above example of bit shifting, the environment in which the processing occurs include the computer (including all the sub-components) and the operator. Essentially the environment is where the process exists during its entire life time. The physical means by which the transition function is applied is the processing mechanism M of the process. In case of a Turing machine for instance, the mechanism consists of head that can read and write symbols on the tape and move the tape left and right one cell at a time. M enables the goal to be achieved by applying the transition rules. The goal, G (I), is defined in terms of the input. For instance, given an input string I=**01001**, we can define a goal to be **G (I)=00011**. The set Σ={0,1} consists of symbols used in the encoding of the input. Notice the above formalism holds for any encoding schema so long the number of symbols used are finite. Γ serves as a storage for the input string. Tape of TM for instance is considered Γ. $Q_0$ is the initial state of the input string along with the environment. For instance, provided the above input, a $Q_0$ of a TM that achieves the goal state can be the input written on the tape, the head pointing at the start of the string (Fig.1). The same notation can be used to represent the goal state (Fig.2).

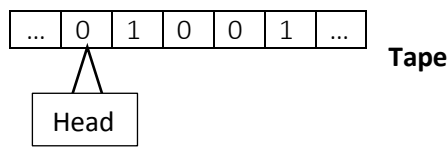

*Figure 1: Start ($Q_0$) configuration of a TM with the input written on the tape and the head reading the first symbol.*

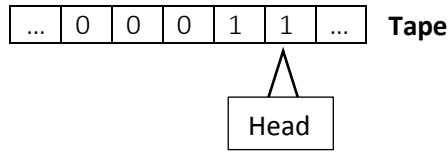

*Figure 2: Final configuration of a TM once goal (G (I)) is achieved.*

Below we show construction of an OP for two real world examples.

**Example 1** – Imagine a person concerned with getting a carton of milk from a nearby store. The process of achieving this goal is an OP where:
- I – You, in the kitchen, without milk (simplified per goal).
- G – You + a gallon of milk in the kitchen.
- O – You or your wife.
- $q_0$ – You in the kitchen, car in the garage, milk at the store.
- Σ – {you, car, care garage, store, milk carto, the road etc.}
- Γ – every physical place these objects can be in (the road, a cart, the trunk etc.)
- Δ – The transition function is a function that constructs the following path:
  
  you → Car garage → car seat → xy intersection → xz intersection → store lot → store entrance → milk aisle → grab milk carton → place on cart → go to register → lot → place milk inside car trunk → car seat → xz intersection → xy intersection → car garage → kitchen
- M – primarily you and your experiences that allow you to navigate from your home to the store and back.

For this example, O initiates the process and checks if the goal state is achieved at every stage.

**Example 2** – To illustrate the scope of OP, consider a more concrete example of bubble sorting integers. An OP with the following components can describe the process of bubble sorting.
- I = {2, 3, 1}
- G – sort I in ascending order– assume each number is written on a piece of paper
- O – You
- $Q_0$ – initial configuration of A on a table top or computer memory
- Σ – every number in A
- Γ – table top, computer memory
- Δ- Scan, compare, swap, stay-same, check goal
- M- a computer or a person

2.2.2 **Definition 2.2:** A *process domain* or *environment* is an environment in which the logical components of the process are encoded using some domain specific language and where the physical components (if needed) are specified.

2.2.3 **Definition 2.2:** An analytic process *(AP)* is *OP* where $G \ != null$ and the start state $Q_0$ is known.

2.2.4 **Definition 2.3:** A computation process *(CP)* is *AP* where M is a mechanical device.

2.2.1 **Thesis:** Any process (physically realistic or not) can be described by *OP*.

2.2.2 **Theorem 2.4:** A running Turing machine M is an OP.
Proof: simply construct an AP as follows:
- I–Input to the TM
- O–you or some automated verification system for verifying goal state.
- G–the expected output of the TM
- Σ –symbols of the device
- Γ–tape of the device
- Δ–rules of transition or simply transition function
- $Q_0$–start state of the input and state of the TM at start
- M–tape, tape head etc …

2.2.3 **DEFINITION 2.4:** Deterministic overarching, deterministic analytic, deterministic computational processes **(DOP, DAP, DCP)** are processes where the transition function at a given stage maps an input state to a set of transition states T where the $|T| = 1$ .

2.2.4 **DEFINITION 2.5**: Non-Deterministic overarching, non-deterministic analytic, non-deterministic computational processes **(NDOP, NDAP, NDCP)** are processes where the transition function at a given stage maps a given state to a set of transition states T where the $|T| = n$ and $n \geq 1$.

2.2.5 **DEFINITION 2.6:** An $i^{th}$ **stage** of a process is a snapshot of the input after applying at least one or more transition rules $i$ times.

2.2.6 **DEFINITION 2.7:** A **state $q_{i_j}$ for a non-deterministic process or $q_i$ for a deterministic variant,** at a given stage *i,* is a distinct configuration (pattern) of the input symbols as recorded by the observer.

2.2.7 **DEFINITION 2.8:** A **step** within a process is the act of transforming state from $q_{i_j}$ to $q_{i_{j+1}}$ or $q_i$ to $q_{i+1}$.

2.2.8 **DEFINITION 2.9:** A **procedure** is a recipe that specifies the steps required to achieve the goal. In other words, it specifies order of operations required to transform input to the desired output.

2.2.9 **DEFINITION 2.10:** *T* is a set of potential transition states for a given input state and $|T| = n \geq 1$.

# 3 ANALYTIC PROCESS (AP)

AP is process that describes the act of achieving a goal starting from a well-defined initial configuration.

**3.1.1** **Definition 3.1**: An analytic process (**AP**) is an 8-Tuple $\langle I, O, G, \Delta, \Sigma, \Gamma, q_0, M \rangle$ where,
- I – input data composed of finite symbols
- G – describes the outcome expected
- O – is an observer system
- $q_0$ – starting configuration of I
- Γ – a structure for storing symbol (grid of cells)
- Δ – a finite rules of transition
- Σ – a finite set of symbols
- M - a mechanism for applying the transition rules

**3.1.2** **Theorem 3.2:** Any computing device undergoes a process that can be described by AP.
**Proof:** simply construct an AP as follows:
- I–input to the mechanical device
- O–you or some automated verification system for verifying goal state.
- G–the expected output of the device
- Σ –symbols of the device
- Γ–hardware of the device
- Δ–rules of the device apply to the input
- $Q_0$–start state of the input and state of the device at the start
- M–the algorithm and the mechanism employed to carry out the algorithm

**Example 1**– AP of a machine that accepts strings containing odd number of 0's and even number of 1's. Let the set of strings containing odd number of 0's and even number of 1's be L and M be a Finite state machine FSA that accepts L (depicted below):

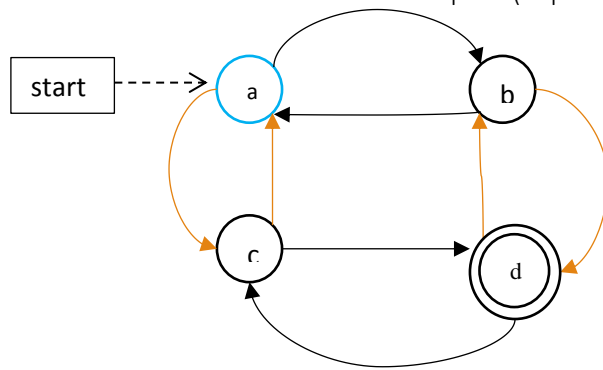

Edges Color code:
- Orange – transition if input symbol is 0
- Black - transition if input symbol is 1
- Blue – start state
- Double circle – accept state

- I – a string w = 001100
- G – does machine M accept. When M reaches the end of w, is it in accept state?
- O – you (or some automated verification system – put together by you)
- Σ – {0,1}
- Γ – a grid of cells for storing 0 and 1

- Δ

| State | 0 | 1 |
|---|---|---|
| a | c | b |
| b | d | a |
| c | a | d |
| d | b | c |

**The process** - Calculate if length of input set modulus cardinality of sigma is 0, if not, it's not valid.
- H – expected iterations of w in Σ =|w| / |Σ|= (6/2) = 3, where |w|is the length of the input and |Σ|is cardinality of the alphabets.
- A – actual iterations of Σ in w = start at 0 such that for every transition state Δ(a,0) up to (|Σ| - 1) has a check if c in w, is also in Σ

1. Place the input string on the grid

    | 0 | 0 | 1 | 1 | 0 | 0 |

2. O presses the start button, applies transition rules
    Start state (a)

    | 0 | 0 | 1 | 1 | 0 | 0 |

3. Δ(a,0) = c, O asks is G achieved? No

    | 0 | 0 | 1 | 1 | 0 | 0 |

4. Δ(c,0) = a, O assesses state. Of available constraints – can A be met by available data remaining NO – go to 9-HALT

    | 0 | 0 | 1 | 1 | 0 | 0 |

5. Δ(a,1) = b, O assesses state

    | 0 | 0 | 1 | 1 | 0 | 0 |

6. Δ(b,1) = a, O assesses state

    | 0 | 0 | 1 | 1 | 0 | 0 |

7. Δ(a,0) = a

    | 0 | 0 | 1 | 1 | 0 | 0 |

8. Δ(a,0) = a, O assess state. End of string: yes, accept state: no.
9. Halt

- **3.1.3** **Definition 3.2:** A deterministic analytic process *(DAP)* is a process in which the transition function $\Delta_{DAP}$ maps an input state at a given stage to a set of *potential transition states T* where the cardinality of $|T| = 1$.
  
  $i.e. \Delta_{DAP}: q_{i-1} \rightarrow q_i$, where $i \geq 1$ is a transformation stage.

- **3.1.4** **Definition 3.3:** A non-deterministic analytic process *(NDAP)* is a process in which the transition function $\Delta_{NDAP}$ maps an input state at a given stage to one or more states drawn from a set of *potential transition states T* where the cardinality of $|T| \geq 1$.
  
  $i.e. \Delta_{NDAP}: q_{i-1} \rightarrow q_{i_j}$, where $i \geq 1$ computation stage and $1 \leq j \leq |T|$.

- **3.1.5** **Lemma 3.1:** DAP can simulate NDAP when $|T_{DAP}| = |T_{NDAP}| = 1$.
  
  Proof: See case 1 of Theorem 3.1

- **3.1.6** **THEOREM 3.1:** *DAP* lacks fundamental capacity in simulating *NDAP* when $|T_{NDAP}| > 1$.
  
  Proof (by contradiction)
  
  Assume **NDAP** can be simulated by **DAP**.
  
  This means we can specify the same goal *G* and input *I* for both. Let the goal be *G* and the input with starting configuration $q_0$ be *I*. Denote the transition function of **DAP** as $\Delta_{DAP}$ and that of **NDAP** as $\Delta_{NDAP}$ and the sets of potential transition states $T_{DAP}$ and $T_{NDAP}$ respectively. From definitions 2.2 and 2.3 we have $|T_{DAP}| \leq |T_{NDAP}|$. Furthermore, let us specify a single observer system, *O*, to simultaneously record transformation of input in each process and to assess goal state. A procedure *p* can be obtained by running the processes as follows:
  
  1. **O** starts **DAP** and **NDAP** with $Q_0$ of *I*
  2. Starting $i = 1$, O applies $\Delta_{DAP}(q_{i-1})$ to generate $q_i$ and $\Delta_{NDAP}(q_{i-1})$ to generate $q_{i_j}$ where $1 \leq j \leq |T_{NDAP}|$. Furthermore, **O** applies $\Delta_{NDAP}$ again to choose one or more values of *j*.
  
     For instance, at $i = 1$ we have $\Delta_{DAP}(q_{1-1}) = \Delta_{DAP}(q_0) = q_1$ and $\Delta_{NDAP}(q_{1-1}) = \Delta_{NDAP}(q_{1_j}) \in \{q_{1_1}, \ldots, q_{1_{|T_{NDAP}|}}\} = T_{NDAP}$
  
  3. **O** checks if **G** is achieved with any procedure
     a. If *G* – return
     b. If not *G* – go to 2
  
  Assume both processes can achieve the goal after **N** computation stages (transformations). Run the two processes as described above.
  
  - Note: here N describes the number of transformations required to achieve the goal. For example, if the goal of a certain process is to sort say {1,4,2,3}, then N=2. This is because after the first transformation I = {1,2,4,3} and after the second we get to the goal state {1,2,3,4}. So, N is the number of distinct states the input can be in including the goal state.

| Stage | DAP | NDAP | |
|---|---|---|---|
| | | Case 1- $\Delta_{NDAP}$ maps to only one of the potential states and is lucky with each guess | Case 2- $\Delta_{NDAP}$ maps to all possible transitions states |
| 1 | 1 | 1 | $|T_1|$ |
| 2 | 2 | 2 | $|T_2|$ |
| 3 | 3 | 3 | $|T_3|$ |
| N | O(N) | O(N) | O ($N*\sum_{k=1}^{N}|T_k|$) |

**Case 1:** Table 1 lists number of transformations applied by each process. Notice NDAP can take the more efficient path (case 1) by employing a mechanism to select one state from the potential states at each stage. Assuming this path takes NDAP to goal state and provided DAP gets to goal state, case 1 proves the equivalence of DAP to NDAP when $|T_{DAP}| = |T_{NDAP}| = 1$ (**Lemma 3.1**). To achieve this NDAP can employ some guessing mechanism that enables it to transition to states, at each stage, that lead to goal state. If the goal is not achieved by NDAP it must try the other potential states (case 2).

**Case 2**: is a more general case where all potential states at every stage are reached. In this case we can clearly see the non-deterministic process requires capacity order of the number of processing stages **N** times the number of potential transition states at every stage **k**. When it reaches goal state NDAP would have reached all the potential states $\sum_{k=1}^{N}|T_k|$. This capacity clearly exceeds the capacity required by the deterministic variant to achieve the same goal. Hence, **DAP** fails to simulate all of **NDAP**. This is a contradiction. **Q.E.D**.

# 4 COMPUTATIONAL PROCESS (CP)

In the early 17th century, before computers became commercially available, the term computer was used to describe a person that performs mathematical calculations (Mahoney, 1988). The meaning was gradually updated to describe programable mechanical machines. In this text we use the term computation process to describe any problem-solving approach that utilizes a programmable mechanical device such as a computer. The term computer is not used as a scientific concept but rather as a tool. The scientific concepts underlying computation are captured by an AP.

### 4.1.1 Definition 4.1: A computation process *(CP)* is *AP* where a mechanical device is used to achieve the goal.

### 4.1.2 Theorem 2.2: Any computing device undergoes a process that can be described by CP.
**Proof:** construct a CP as follows:

- **I**–input to the computing device
- **O**–you or some automated verification system for verifying goal state
- **G**–the expected output of the device
- **Σ** –symbols encoding input to the device
- **Γ**–hardware of the device (RAM etc)

- Δ–rules of the device apply to the input to generate desired output
- $Q_0$–start state of the input and state of the device at the start
- $M$–the algorithm and the mechanism employed to carry out the algorithm (such as CPU, hard drive)

### 4.1.3 Definition 4.2:
A deterministic computational process *(DCP)* is a process in which the transition function $\Delta_{DCP}$ maps an input state at a given stage to a set of *potential transition states T* where $|T| = 1$.

$i.e. \Delta_{DCP}: q_{i-1} \rightarrow q_i$ , where *i* ≥ 1 computation stage.

### *4.1.4* Definition 4.3:
A non-deterministic computational process *(NDCP)* is a process in which the transition function $\Delta_{NDCP}$ <u>maps</u> an input state at a given stage to one or more states drawn from a set of *potential transition states T* where the cardinality of $|T| = n \geq 1$.

$i.e. \Delta_{NDCP}: q_{i-1} \rightarrow q_{i_j}$ , where *n* ≥ 1 computation stage and $1 \leq j \leq |T|$.

### 4.1.5 Lemma 4.1:
*DCP* is a bounded *NDCP* when $|T_{DCP}| = |T_{NDCP}|$ .

Proof:
- We know DAP = NDAP when $|T_{DAP}| = |T_{NDAP}|$ **(Lemma** 3.1).
- By Definition 4.1 CP is AP performed by a computer.
- Then it follows that **DCP = NDCP** when $|T_{DCP}| = |T_{NDCP}|$ . Q.E.D

### 4.1.6 Theorem 4.1:
Deterministic-computational process *(DCP)* lacks fundamental capacity in simulating a non-deterministic-computational process *(NDCP).* In other words, implementation of a given non-deterministic-computational process inherently requires dimensions that exceed the deterministic variant.

Proof:
- CP is AP where the mechanism used to transform input is a mechanical computing device. Clearly AP is stronger than CP by the fact that CP has physical constraints imposes as a result of using mechanical device.
- From Theorem 3.1 we know DAP is unable to simulate NDAP.
- It follows that DCP is unable to simulate NDCP. Q.E.D

# 5 PROPERTIES

(CLOSURE) DEFINITION 5.1: HYBRID PROCESS - is a process in which both deterministic and non-deterministic processes are used to achieve the goal state.

- 5.1.1 **Definition 5.2: D-Hybrid process** – combining two or more deterministic processes results in a deterministic process.
- 5.1.2 **Definition 5.3: DND-Hybrid process** – Combining one or more deterministic processes with one or more non-deterministic processes results in an overall non-deterministic process.
- 5.1.3 **Definition 5.4: ND-Hybrid process** – When two or more non-deterministic processes are combined the result is non-deterministic.

# 6 OCM AND TURING MACHINE

- 6.1.1 **Definition 6.1**. A Turing Machine (**TM**) is a 7-tuple T$M = \langle Q', \Sigma', \Gamma', \Delta', q'_0, q'_{accept}, q'_{reject} \rangle$
  - $Q'$ – a finite set of transition states
  - $\Sigma'$ – a finite set of symbols (input alphabets)
  - $\Gamma'$ – a finite set of tape alphabets
  - $\Delta'$ – transition function
  - $q'_0$ – initial state
  - $q'_{accept}$ - accepting state
  - $q'_{reject}$ - rejecting state

- 6.1.2 **Theorem 6.1:** Any Turing machine **M1** running to determine membership *of a language L* is an AP.
  **Proof:**
  Simply construct an AP instance with
  - **O** – some automated verification system or you
  - **M** = M1's tape, head etc
  - **G** = Does M1 accept the language L? Does **M1** accept *w,* $\forall w \in \Sigma$?
  - $\Sigma = \Sigma'$
  - $\Gamma = \Gamma'$
  - $\Delta = \Delta'$

- 6.1.3 **Corollary 6.1:** A **Turing process (TP)** is an AP where a Turing machine is used to achieve the goal.
  **Proof:** Theorem 6.1

- 6.1.4 **Definition 6.2:** A **Deterministic Turing Process (DTP)** is an analytic process where a deterministic Turing machine is used.

- 6.1.5 **Definition 6.3:** A **Non-Deterministic Turing Process (NDTP)** is an analytic process where a non-deterministic Turing machine is used.

- 6.1.6 **Theorem 6.2:** A **DTP** running to determine membership *of a language L* is a **DAP**.
  **Proof:**
  This is due to the fact that the transition function of *DTP* maps every symbol to one state.
  - We know DTP is TP (Definition 6.2)
  - DTP is AP since all TPs are AP (Theorem 6.1)
  - DTP is DAP. By definition the transition function $\Delta'$ of DTP maps a given symbol to one state. This is equivalent to a DAP. **Q.E.D**

- 6.1.7 **Theorem 6.3:** A **NDTP** running to determine membership *of a language* **L** is a **NDAP**.
  **Proof:**
  This is due to the fact that the transition function of NDTP maps every symbol to multiple states.
  - We know NDTP is a TP (Definition 6.3)
  - NDTP is AP since all TPs are AP (Theorem 6.1)
  - NDTP is NDAP. By definition the transition function $\Delta'$ of NDTP maps a given symbol to multiple states. This is equivalent to a NDAP. **Q.E.D**

- 6.1.8 **Theorem 6.4:** DTP lacks fundamental capacity to simulate all of NDTP.
  **Proof:**
  - We know DTP is DAP (Theorem 6.2) and NDTP is NDAP (Theorem 6.3)
  - DAP lack fundamental capacity to simulate NDAP (Theorem 3.1)
  - Then it follows that DTP fails to simulate NDTP. **Q.E.D.**

# 7 OCM AND P VS NP

In this section we attempt to shed light on the long standing unresolved **P vs NP** problem (Cook, 1998) using the OCM framework. Informally, the problem is concerned with whether every language accepted by some nondeterministic algorithm in polynomial time is also accepted by some (deterministic) algorithm in polynomial time. Hence, the class **P** represents all problems solvable by deterministic algorithms in polynomial time while the problems residing in the class **NP** are those that can be solved by non-deterministic algorithms in polynomial time. If one adheres to defining NP in terms of polynomial time verifiability, then this proof can be used as additional tool to separate the two classes. However, we believe verifiability of a solution to NP or any problem has nothing to offer regarding the solvability of the problem (this is stated as an Axiom and we do not have concrete proof or an approach at the moment). Our intuition is that the two classes are related to fundamental aspects of nature and offer the following view, (driven from Theorem 3.1):

> *'Deterministic algorithm lacks <u>fundamental capacity</u> to fully simulate a non-deterministic variant'*

In practice, a deterministic algorithm can approximate a non-deterministic variant to some degree. However, deterministic and non-deterministic approaches to a decision problem are inherently different. Particularly, a non-deterministic approach requires dimensions of greater magnitude compared to a deterministic variant, making the deterministic inferior in its capabilities. As you have noted, the above statement makes no mention of space, time bounds or verifiability. As stated above, verifiability can be used as an additional criterion for grouping NP problems, but we do not think it has anything to offer regarding the solutions. The space and time constraints can be addressed using CP that assumes physical constraints. Our primary goal here is to convey our view on determinism and non-determinism processes in broad sense and their implication to the P vs NP question.

### 7.1.1 Definition 7.1 (The class P).
The class **P** is the set of problems solvable by deterministic Turing Process (DTP).

$$P = \{L \mid L = L(M) \text{ for some DTP } M\}$$

### 7.1.2 Definition 7.2 (The class NP).
The class **NP** is the set of problems solvable by non-deterministic Turing Process (NDTP).

$$NP = \{L \mid L = L(NM) \text{ for some NDTP } NM\}$$

### 7.1.3 Lemma 7.1:
$l$ is a DAP $\forall l \in P$. In other words, a Deterministic Turing Process running to solve any problem in *P* is a DAP.

**Proof:**

- We know a **DTP** is a **DAP** (Theorem 6.2)
- The class **P** is exactly the class of problems solvable by **DTP** (by construction, Definition 7.1)
- Thus, it follows that $l$ **is a DAP** $\forall l \in P$. **Q.E.D**

### 7.1.4 Lemma 7.2:
$l$ is a NDAP $\forall l \in NP$. In other words, a Deterministic Turing Process running to solve any problem in N*P* is a *NDAP*.

**Proof:**
- We know a **NDTP** is a **NDAP** (Theorem 6.3)
- The class **NP** is exactly the class of problems solvable by **NDTP** (by construction, Definition 7.2)
- Thus, it follows that $l$ **is a NDAP** $\forall l \in NP$. **Q.E.D**

### 7.1.5 Axiom 7.1:
Verifiability of a solution **s** to a problem **p** offers no information regarding a process required to solve that problem.

### 7.1.6 Theorem 7.1: $P \neq NP$

Proof (by contradiction):
Assume P = NP → **P** ⊂ **NP** and **NP** ⊂ **P**

→ $P \subset NP$
- $l$ is a **DAP** $\forall l \in P$ (Lemma 7.1)
- $ln$ is a **NDAP** $\forall ln \in NP$. (Lemma 7.2)
- We know all of **DAP** can be simulated by **NDAP** (Lemma 3.1) ➔ $P \subset NP$
    - **P** is contained in **NP** since one can find a non-deterministic algorithm for every deterministic algorithm.

← $NP \subset P$?
- Let $I_i \in NP$ be a problem solvable by some non-deterministic algorithm $A_i$ for $0 < i < |NP|$. Where $|NP|$ is the cardinality of the set NP or the number of problems classified as NP.
- Each $A_i$ is a **NDAP**. (by Definition of the class NP)
- We know $A_i$ cannot be simulated by any deterministic algorithm for $0 < i < |NP|$. (Theorem 3.1)
- Therefore $NP \not\subset P$. This is a contradiction.
- Hence $P \neq NP$. **Q.E.D**

# 8   CONCLUSION

We conclude by highlighting the philosophical implications of this framework and the above proof on computation and mathematics. When algorithm designers are engaged in designing a procedure to solve a problem, they are equivalently engaged in weeding out paths that do not lead to a solution. When a solution is specified in the form of a procedure or algorithm, its then verified for different instances of the problem. In general, if the solution holds for all instances, this is considered an exact solution, otherwise it's wrong or an approximation. In computation complexity, the field that studies classes of problems according to their inherent difficulty, many problems have been classified into several classes according to the resources and time required to implement their solution. We argue that this is equivalent to the study of resources required to implement a computational process. As a result, its concerned with the fundamental limits of nature as oppose to the nature of the problems themselves. In addition, we argue that the existence of evolution forces us to consider the designer as an integral part of the problems. To elaborate, let us consider the following setup while keeping the distinction between determinism and non-determinism in mind.

Imagine a driver and bicyclist who plan to arrive at city B starting from A (Fig. 3). Both cities are connected by city roads and\or bike lanes. The operators must drive\ride pass other cities to arrive at the destination. For practical completion, there also exist dead-ends. *Most cities have car roads but not necessarily bike lanes* while others have exclusive lanes alongside the main roads. The problem states: arrive at B starting from A. Note here we are not concerned with efficiency. *Also assume the problem is posed to the driver and biker who are not aware of the paths except for the signs that indicate bike or car friendliness*. The question is not posed to the reader.

Biker starts at A and choses the next city reachable via bike lane. For instance, the city at position (0,3). Next, they decide to ride to (1,2) then (3,1) followed by (4,3) before arriving at the destination. When put together the biker's options are:

$$A \to (0,3) \to (1,2) \to (3,1) \to (4,3) \to B$$

$$etc \ldots$$

The biker has a total of 8 possibilities. Similarly, the driver can arrive at the following or any of the other 1496 practical procedures by engaging in a similar discerning approach:

$$A \to (0,2) \to (1,2) \to (3,2) \to (4,2) \to B$$

$$A \to (0,0) \to (1,2) \to (3,2) \to (4,2) \to B$$

$$A \to (0,3) \to (1,2) \to (3,2) \to (4,2) \to B$$

$$A \to (0,4) \to (1,2) \to (3,2) \to (4,2) \to B$$

$$etc \ldots$$

The difference between the two approaches is not mere difference between a bike and car. It's the very setup of the problem including the roads and lanes, the possible number of paths, the persons etc… The bike's options to achieve the goal are significantly fewer compared to the driver's. The driver has 1496 more potential paths.  If the biker had a car or could drive, he would have the same number of potential paths. However due to the nature of the setup the biker has limited options. Similarly, the driver could be out of lack if her car broke down at the start in which case she will resort to plan B of riding a bike. These limitations are not only because of the vehicles but also the pavements and roads. Had the city planers decided to pave bike lane along side of every car road, the biker would have had the same number of paths.

Analogues to this setup, solving any problem, including mathematical, involve constraints the solver must deal with to achieve a solution. Many of these constraints are naturally imposed. These can be lack of proper tools and knowledge just like the biker is out of luck for not having a car. They can also be due to the very nature of reality. Just like we are unable to drive faster than the speed of light or the limit on the number of transistors we can fit within a CPU without melting it. This notion is mostly relevant to computable problems. Specifically, to the complexity classes encountered in computational complexity theory. For instance, the class P is a set of efficiently computable problems. We argue a more precise definition for this is a set of efficiently computable problems given our existing tools and our current understanding of the nature of reality. Similarly, other complexity classes should consider similar observation. Those problems that do not have efficient algorithm are not necessarily inherently hard problems. Just like the biker for having only a bike, these problems are not inherently hard but rather hard given the tools we possess, and the constraints imposed due to nature. Thus, one must not only consider the limits of computation imposed by nature but also our interpretation of the nature of reality. After all the world we live in is not static. Many of our achievements are attributed to evolution. Just like the biker can save up and get himself a car and expand his options, the problems that are considered hard now may not be in the near future. Considering them inherently hard is only reminiscent of our limitations and the framework we operate from.

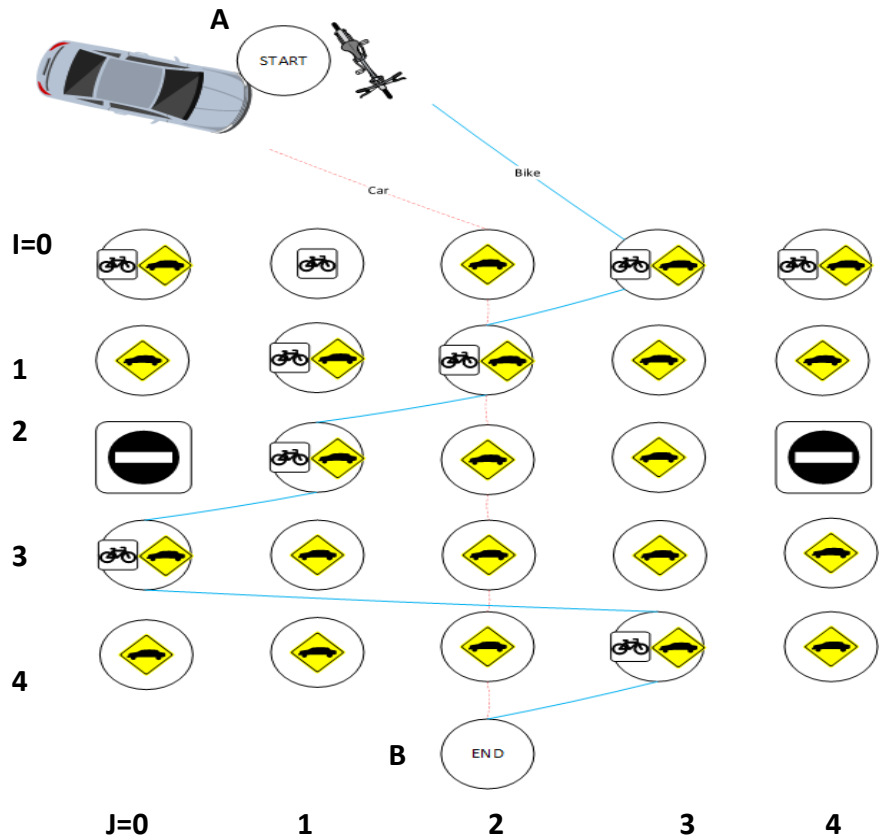

*Figure 3:*

The same illustration can be employed to grasp the difference between a deterministic and non-deterministic algorithm. For simplicity, let us assume the biker only has a single path to achieve his goal. One can easily see the process undertaken by the biker is a deterministic process. While the driver's is non-deterministic. Similar arguments can be invoked for why these two processes are different. The biker is forced to be a deterministic process due to the constraints of the setup. The driver however has the luxury to choose form all the different paths because there happen to be more car roads than bike lanes. Similarly, many problems we encounter in many fields can potentially be solved by some non-deterministic approach while few can be using deterministic paths. Our observation is that its due to the very nature of the problems that the designer's approaches are deterministic or non-deterministic. This is to say non-deterministic processes are fundamental with deterministic being a special case.

Another point we want to touch on with this illustration is notion of efficiency. Deterministic algorithm is more efficient as it requires less resources than a non-deterministic variant. The biker's path requires few number of parameters to encode than the drivers'. This is due to the fact that the biker's path is known or prespecified. For instance, the single path (blue in Fig. 3) taken by the biker can be encoded using the string $A \rightarrow (0,3) \rightarrow (1,2) \rightarrow (3,1) \rightarrow (4,3) \rightarrow B\ (\#)$ or using similar representations. The number of symbols required are significantly fewer than that of a non-deterministic procedure. For the above example, one needs approximately 180 * # more number of symbols to represent the non-deterministic procedure. This is due to the fact that non-deterministic approaches consist of more potential paths. First one must have knowledge of all the potential paths (or approaches to resolving the paths) that solve the given problem beforehand then one must implement all the potential paths. We say 'must implement all potential paths' because if one reduces the number of paths by any means then we are collapsing the solution back to deterministic. This is the underlying reason why deterministic fails to simulate non-deterministic algorithm.